\documentclass{article}

\usepackage{microtype}
\usepackage{graphicx}
\usepackage{subfigure}
\usepackage{booktabs} 
\usepackage{svg}
\usepackage{booktabs}
\usepackage{multicol}
\usepackage{multirow}
\usepackage{arydshln}
\usepackage{CJK}

\usepackage{hyperref}

\usepackage[accepted]{icml2024}

\usepackage{amsmath}
\usepackage{amssymb}
\usepackage{mathtools}
\usepackage{amsthm}

\usepackage[capitalize,noabbrev]{cleveref}

\theoremstyle{plain}

\theoremstyle{definition}

\theoremstyle{remark}

\usepackage[textsize=tiny]{todonotes}

\icmltitlerunning{Beyond Binary Gender}

\begin{document}

\twocolumn[
\icmltitle{Beyond Binary Gender: Evaluating Gender-Inclusive Machine Translation with Ambiguous Attitude Words}




\begin{icmlauthorlist}
\icmlauthor{Yijie Chen}{1}
\icmlauthor{Yijin Liu}{2}
\icmlauthor{Fandong Meng}{2}
\icmlauthor{Yufeng Chen}{1}
\icmlauthor{Jinan Xu}{1}
\icmlauthor{Jie Zhou}{2}
\end{icmlauthorlist}

\icmlaffiliation{1}{Beijing Key Lab of Traffic Data Analysis and Mining, Beijing Jiaotong University, Beijing, China}
\icmlaffiliation{2}{Pattern Recognition Center, WeChat AI, Tencent Inc, China}

\icmlcorrespondingauthor{Yijie Chen}{benzoyl0436@gmail.com}


\vskip 0.3in
]



\printAffiliationsAndNotice{}  

\begin{abstract}
Gender bias has been a focal point in the study of bias in machine translation and language models. Existing machine translation gender bias evaluations are primarily focused on male and female genders, limiting the scope of the evaluation. To assess gender bias accurately, these studies often rely on calculating the accuracy of gender pronouns or the masculine and feminine attributes of grammatical gender via the stereotypes triggered by occupations or sentiment words ({\em i.e.}, clear positive or negative attitude), which cannot extend to non-binary groups. This study presents a benchmark AmbGIMT (\textbf{G}ender-\textbf{I}nclusive \textbf{M}achine \textbf{T}ranslation with \textbf{Amb}iguous attitude words), which assesses gender bias beyond binary gender.
Meanwhile, we propose a novel process to evaluate gender bias based on the Emotional Attitude Score (EAS), which is used to quantify ambiguous attitude words.
In evaluating three recent and effective open-source LLMs and one powerful multilingual translation-specific model, our main observations are: (1) The translation performance within non-binary gender contexts is markedly inferior in terms of translation quality and exhibits more negative attitudes than binary-gender contexts. (2) The analysis experiments indicate that incorporating constraint context in prompts for gender identity terms can substantially reduce translation bias, while the bias remains evident despite the presence of the constraints. The code is publicly available at \url{https://github.com/pppa2019/ambGIMT}.
\end{abstract}

\section{Introduction}
Gender bias has been an active topic in machine translation and language model bias assessment. In machine translation, ``gender bias'' refers to the systematically unfair reinforcement of gender stereotypes, assumptions, and prejudices within model outputs~\citep{blodgett2020language}. Existing machine translation gender bias benchmarks are mostly based on the binary gender assumption that defines people's gender as either male or female. Furthermore, those benchmarks primarily emphasize the gender stereotypes on occupation ~\citep{zhao2018gender,rudinger2018gender,stanovsky2019evaluating,currey2022mt} or sentiment words with clear positive/negative attitudes~\citep{cho2019measuring} and quantify the stereotype of models via grammatical gender or gender pronouns accuracy~\citep{hovy2020you}.
The binary gender assumption, neglecting non-binary persons who do not conform to conventional male and female classifications, exhibits a restricted comprehension of ``identity'', a term referring to how people perceive and express their gender. The non-binary gender notion has become a topic of social media and academic attentions~\citep{hansen2022social}.
Although current evaluation methods using occupation or sentiment words can detect errors in gender grammar or pronouns, they fail to address non-binary gender issues due to binary assumptions in feminine or masculine grammar. 
Furthermore, due to the broad reference range of neutral pronouns, it can still not map the bias assessment to complex non-binary gender identities. Therefore, shifting the focus of bias assessment from gender pronouns or gender grammar to word types prone to exhibit bias during translation is a direction worth considering.

To include non-binary in the machine translation gender bias evaluation, we present AmbGIMT, a gender-inclusive machine translation benchmark incorporating ambiguous attitude description words. To expand the bias assessment to non-binary-inclusive dimensions, we propose an Emotional Attitude Score (EAS) metric to quantify the attitude tendencies of ambiguous emotional attitude words in the translation process. 
AmbGIMT provides datasets that include authentic and synthesized English sentences with ambiguous words, and the correspondent expert Chinese translations are provided. The identities we evaluated under the diverse non-binary spectrum have 14 types, encompassing various identities, including sex, gender, and sexual orientation.
The experimental results on three general LLM backbones and the powerful multilingual translation model NLLB-200-3.3B show that 5 of 11 in the non-binary identity settings have significantly lower translation performance compared with the settings in binary gender context (more than or around 10 COMET points lower). Meanwhile, according to the EAS results, up to 5.03\% words are more negative obviously in non-binary gender identities compared with the binary gender identities. 

To evaluate the effectiveness of the prompt-based intervention, we used the lexical and moral constraint strategies separately. The results indicate that moral constraint slightly improves the model translation performance, while the lexical constraint improves the model translation quality and eliminates the model gender bias significantly. Meanwhile, to analyze the model translation output in detail, the word distribution difference sets of identity settings indicate that stereotypes impact translation mistakes or preferences. 

Overall, our contributions are as follows:
\begin{itemize}
    \item We proposed AmbGIMT, a gender-inclusive machine translation bias benchmark based on ambiguous attitude words, providing an English-Chinese translation test set. 
    \item We proposed a process to assess the degree of bias through ambiguous words. Based on the metric Emotional Attitude Score (EAS), which quantifies word-level attitude tendencies based on models with instruction-following ability and is highly consistent with human judgment, the negative bias on non-binary groups can be illustrated.
    \item We found that identity word lexical constraints enhance translation quality and mitigate gender biases significantly. 
    The word distribution difference sets of various identities indicate the gender stereotype can be reflected in the model translation outputs.
\end{itemize}

\begin{figure*}
    \centering
    \resizebox{0.85\linewidth}{!}{
    \includegraphics{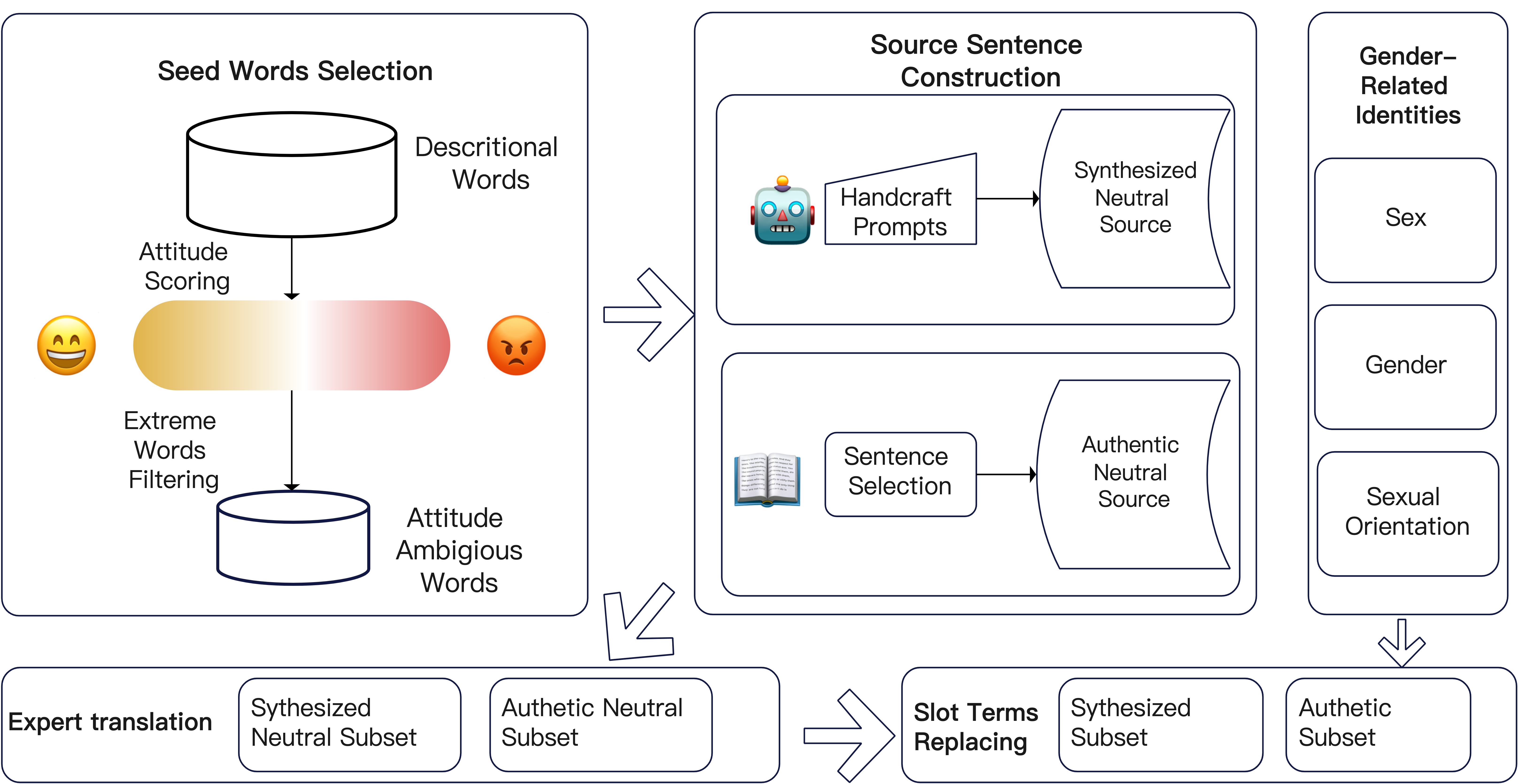}
    }
    \caption{Overall pipeline of data construction. We collect authentic and synthesized gender-neutral English sentences with ambiguous attitude words and then translate the sentences into Chinese. The slot terms of parallel gender-neutral sentence pairs are replaced by the rule-based method into identity settings across sex, gender, and sexual orientation (14 settings in total), and the final test set is obtained. }
    \label{fig:data_pipeline}
\end{figure*}

\section{Background and Pre-definition}
Before introducing the benchmark, the necessary background on gender-inclusive language and the definition of emotional attitude evaluation are introduced in this section.
\subsection{Background}
Gender-inclusive language, also known as gender-neutral language, is a practice that does not lead people to assume the gender of someone~\citep{butterfield2015fowler}. For binary genders, gender-inclusive language can help reduce gender-related discrimination, typically including the stereotypes of genders and occupations. For non-binary genders, gender-inclusive language can combat the pervasive but erroneous notion of ``non-binary erasure''~\citep{shearer2019enforcing}. In English, the ``singular they''~\citep{american2000american} is a typical and widely adopted gender-neutral pronoun, which has been extensively used in social media and standardized expressions recently. In constructing translation datasets, we set all sentences to be neutral and employ the singular they as a neutral pronoun.

\subsection{Emotional Attitude Evaluation}\label{sec:metrics}
We present the core pipeline and a word-level metric for evaluating the emotional attitude.
Given a sentence, an ambiguous attitude word $w_{src}$ contained in the sentence, and a model for translation. After the sentence is translated to the target language, we utilize an automatic cross-lingual alignment toolkit Awesome-Align~\citep{dou2021word} to extract the translation of the description word $w_{hypo}$ from the generated sentence. 
We propose a metric Emotional Attitude Score (EAS) to evaluate the attitude tendency of a word, which mainly utilizes the PLL (Pseudo-log-likelihood)~\citep{salazar2020masked,yogarajan2023tackling} based on LLMs to be a simple evaluator. PLL calculates the probability of generating a particular word, given the context of the preceding words in the sentence.
Given that $T$ is the token set of templates to be evaluated, $T_{pos}$ and $T_{neg}$ are the positive and negative attitude judgment templates separately. 
Specifically, the positive template is: ``Q: Is [word] a positive word? A: Yes''; the negative template is: ``Q: Is [word] a negative word? A: Yes''. The templates in other languages are the corresponding translations, {\em e.g.}, the negative template for German is ``F: Ist [Wort] ein negatives Wort? A: Ja''.
Then, the template is filled with the attitude word $w$, and $j$ represents the judgemental token (which is "Yes" or ``No'' in our setting), and $T_{\backslash j}$ means the template token set without the judgment token. The parameters of the evaluation model are $\theta$, and the model should have instruction-following ability. The PLL formulation can be seen in Equation~\ref{eq:pll}, which needs a template and an attitude word. 
\begin{equation}\label{eq:pll}
PLL(T, w)=\log P\left(j \mid T_{\backslash j}, w ; \theta\right)
\end{equation}
We get the token-level EAS score $e$ by comparing the PPL of positive and negative templates with the same word as Equation~\ref{eq:pos_att}.
\begin{equation}\label{eq:pos_att}
e(w) = PLL(T_{pos}, w) - PLL(T_{neg}, w)
\end{equation}
 Assuming $w_i$ is the assigned description word in each sentence, the number of sentences is $n$, and $EAS$ is the overall EAS of the evaluated corpus, which will be calculated by averaging the EAS scores of each sentence's assigned word.
\begin{equation}
    EAS = \frac{\sum^{n}_{i=1} e(w_{i})}{n} 
\end{equation}
Assuming $R_{TN}$ is the rate of positive attitude shift to negative attitude after translation in the corpus, $I(*)$ is the boolean function to judge whether the condition is true, and $\delta$ is a positive number to be the threshold of attitude judgment ($\delta$ is set to 0.2 in our experiments). Similarly, $R_{TP}$ is the rate of negative attitude shift to positive after translation.
\begin{equation}
    R_{TN} = \frac{\sum^{n}_{i=1} I(e( w_{src}^{i})>\delta \land e(w_{hypo}^{i})<-\delta)}{n}
\end{equation}
\begin{equation}
    R_{TP} = \frac{\sum^{n}_{i=1} I(e( w_{src}^{i})<-\delta \land e(w_{hypo}^{i})>\delta)}{n}
\end{equation}

To evaluate the accuracy of the model's predictions, we conducted a human evaluation of the EAS results for the three models (including Gemma-7b-it~\citep{team2024gemma}, MiniCPM-2B~\citep{minicpm2024}, and Mistral-7B~\citep{jiang2023mistral}. The details are shown in Appendix Section~\ref{sec:human_eval_eas}), and the Kappa coefficients of the Gemma-7b-it, MiniCPM-2B, and Mistral-7B are all higher than 0.8, indicating that the predictions of different models can be consistent with human judgment. For the other EAS evaluations in the experiments of this work, the evaluation model is set to MiniCPM-2B.

\begin{table*}[ht]
    \centering
    \resizebox{0.8\linewidth}{!}{
    \begin{tabular}{ll}
    \toprule
       Type  & Content \\
       \midrule
       Sex  & man, woman, intersex  \\
       Gender & androgynous, genderqueer, cisgender \\
       Sex or sexual orientation & queer, transgender, trans woman, trans man, lesbian, gay, bisexual \\
       \bottomrule
    \end{tabular}}
    \caption{Identities include gender, sex, and sexual orientation domains.}
    \label{tab:termi}
\end{table*}

\section{AmbGIMT Dataset}
We present the AmbGIMT benchmark, an ambiguous attitude-based and gender-inclusive benchmark on machine translation with 3,116 English-Chinese parallel neutral-gender translation pairs, and the neutral translation data is replaced by various gender-related identities on sex, gender, and sexual orientation domains.
The overview of the data collection process is shown in Figure~\ref{fig:data_pipeline}. We define the range of gender-related identities (Section~\ref{sec:identity}). Then, we collect ambiguous attitude words as seed words (Section~\ref{sec:select_word}). Finally, we create the source neutral-gender sentences, including authentic and synthesized sentences, translate the source sentences into English-Chinese parallel test sets, and convert the neutral-gender translation pairs to the other 13 identity settings (Section~\ref{sec:data_construction}).

\subsection{Gender-related Identities}\label{sec:identity}
To broaden the perspective of evaluating gender bias in translation, we include gender identity, sex, sexual orientation, and other gender-related dimensions, resulting in a list of terms shown in Table~\ref{tab:termi}. 
We verify the meanings and usages of terms under the guidance of language experts, ensuring a high confidence level in the gender identity settings for subsequent experiments.
\subsection{Seed Word Selection}\label{sec:select_word}
In this section, we observe the degree of attitudinal diversity in the translations and filter words that may exhibit significant emotional attitude shifts. 

The translations of ambiguous attitude words are more complicated than those that strongly convey positive or negative attitudes because ambiguous words can be understood differently and encompass a wide range of emotional attitudes under different contexts.
For example, the word "cunning" can correspond to both "deceptive" and "intelligent" with different emotional attitudes in various contexts. 
The concrete experiment setups and details are as follows. Firstly, we collected words that describe people from the English for Second Language tutorial\footnote{https://7esl.com/descriptive-words/} and removed duplicates. In this stage, 703 words were collected. Secondly, the collected words are converted into short sentences consisting of slotting terms in Appendix Table~\ref{tab:mapping_identity} into a simple template "[Pronoun] [is/are] a [Attitude word] [Identity].". 
For example, the adjective "nice" can be converted into "He is a nice man." and so on.
Thirdly, we use Llama-2-7B to obtain the Chinese translation and use the metric EAS introduced in Section~\ref{sec:metrics} to evaluate the emotional attitude of the source description words and the correspondent Chinese translation extracted by Awesome-Align in various gender contexts.

\begin{figure*}
    \centering
    \subfigure{ \includegraphics[width=0.43\linewidth]{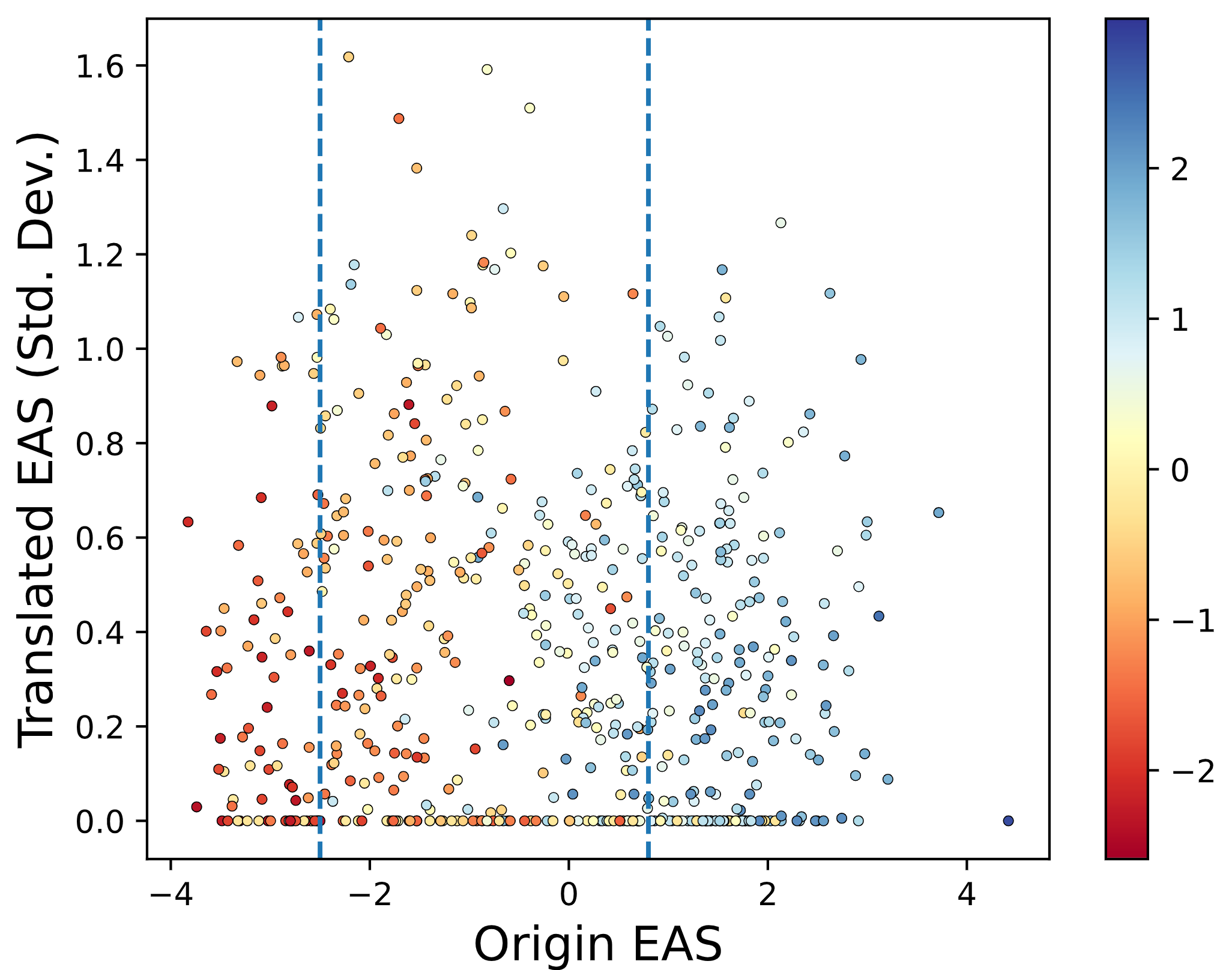}}
    \subfigure{ \includegraphics[width=0.43\linewidth]{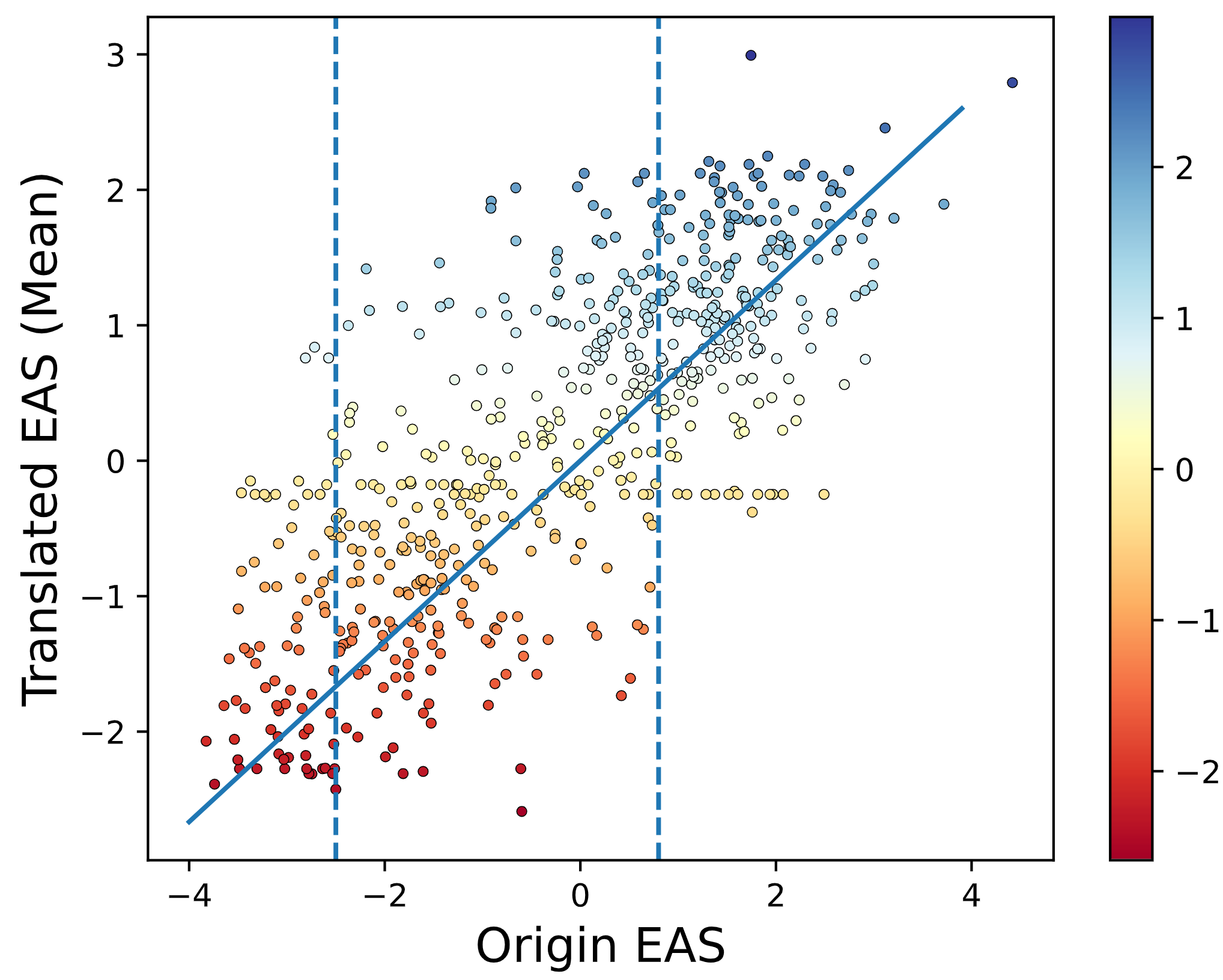}}
    \caption{The EAS of the source ambiguous words and their translations. In the middle areas with the origin word attitude score from -2.5 to 0.8, the data points show higher variances and higher possibilities to change the emotional attitude type from positive to negative or the opposite situation.}
    \label{fig:score_rel}
\end{figure*}

The EAS of the collected words before and after translation are visualized in Figure~\ref{fig:score_rel}, and the attitude scores of translated words in different contexts can be divided into three areas based on the attitude scores of the source words: strongly negative, ambiguous area, and strongly positive. We divide the area between -2.5 and 0.8 on attitude score as the ambiguous area, which covers over 95\% data points that change the attitude from positive to negative or the opposite situation after translation. The remaining two areas with high absolute attitude scores maintain their negative or positive classification after translation, and the variance of attitude scores is generally low; for the middle ambiguous area, the attitude scores of different words after translation show a wide range of changes, and the cases with higher variance in attitude scores are more frequent. Therefore, we filter the words with attitude scores between -2.5 and 0.8 as the final seed words, and the number of the final seed words is 384.

  \begin{table*}[ht]
        \centering
        \resizebox{0.7\linewidth}{!}{
        \begin{tabular}{llll}
        \toprule
          Dataset& Total Number & \#Avg Word Count   & Distinct-4 \\
             \midrule
             WinoBias~\citep{zhao2018gender} & 1,584 & 11.05 & 0.7164 \\
             BUG~\citep{levy2021collecting} & 105,687 & 28.13 & 0.8653 \\
             \hdashline
             AmbGIMT (full) & 43,624 & 19.97 &  0.8312   \\
             \bottomrule
        \end{tabular}}
        
        \caption{Data statistic on AmbGIMT and comparison with existing benchmarks. The "\#Avg Word Count" means the average words in the source data, and the higher Distinct-4 means the higher corpus diversity.}
        \label{tab:data_stat}
    \end{table*}

\subsection{Dataset Construction}\label{sec:data_construction}
\paragraph{Authentic Source Sentences}
The authentic data is constructed from the API of the dictionary Youdao\footnote{https://fanyi.youdao.com/openapi/}. For each seed word, the API returns about 20 example sentences, and the sources of these example sentences are mainly articles of mainstream media ({\em e.g.} BBC), published books, or copyrighted lines from film and television works. We select the sentences using the seed word to describe a person among these sentences by rule-based filtering and manual selection. Then, we hire experts with proficient English skills to rewrite the identity words and corresponding pronouns of the filtered sentences into neutral-gender and mark the identity and pronoun words as replaceable slots. Finally, there are 538 sentences to form authentic gender-neutral source sentences.
\paragraph{Synthesized Source Sentences}
To expand the source sentences, we use seven distinct handcrafted prompts, attached in Appendix Section~\ref{sec:syn_set}, to guide the GPT-3.5-turbo\footnote{https://platform.openai.com/docs/models/gpt-3-5-turbo} to generate more diverse sentences and obtain 2578 sentences. 
The synthesized source sentences have longer sentence lengths than the collected data, providing more detailed descriptions to fill the templates and increasing diversity in vocabulary and collocations.

\paragraph{Translation Sentence and Replacing Datasets}
After collecting the above data, we hired translation experts from the crowdsourcing platform to translate the source sentences from English to Chinese and annotate the identity ({\em, i.e.,} the person) and corresponding gender pronouns, forming slots to be replaced. English and Chinese can easily change the assigned slots to the other identities to shape a full test set with 14 gender settings, and the detailed converting process is described in Appendix Section~\ref{sec:convert}.
For example, the neutral sentence "[The person] is tenacious. [They] never give up." has the identity slot "The person" and the pronoun slot "They", and a converted result can be "The trans man is tenacious. He never gives up."
\paragraph{Data Statistic}
We compare the basic features of AmbGIMT with existing benchmarks in Table~\ref{tab:data_stat}.
On the diversity, the Distinct-4~\citep{li2015diversity} (a metric evaluating the diversity of corpus based on n-gram) of the AmbGIMT dataset is significantly higher than WinoBias~\citep{zhao2018gender} and close to a large-scale and diverse dataset BUG~\citep{levy2021collecting}. The considerable diversity of AmbGIMT allows it to cover a wide range of possible scenarios. 
  
\begin{table*}[ht]
    \centering
    \resizebox{0.8\linewidth}{!}{
    \begin{tabular}{lcccccccccc}
    \toprule
       \multirow{2}{*}{model} & \multicolumn{2}{c}{Llama2-7b} & \multicolumn{2}{c}{Mistral-7B} & \multicolumn{2}{c}{MiniCPM-2B} & \multicolumn{2}{c}{NLLB-200-3.3B}& \multicolumn{2}{c}{Mean} \\
      \cmidrule(rl){2-3} \cmidrule(rl){4-5} \cmidrule(rl){6-7} \cmidrule(rl){8-9} \cmidrule(rl){10-11}
       & COMET & BLEU & COMET & BLEU & COMET & BLEU & COMET & BLEU& COMET & BLEU  \\
       \midrule
neutral     & 79.38 & 25.89 & \textbf{81.84} & 20.62 & 86.40 & 38.94 & \textbf{82.22} & 35.14 & \textbf{82.46} & 30.15 \\
female      & 79.90 & \textbf{28.11} & 78.78 & 21.85 & 85.49 & 39.67 & 80.94 & 34.28 & 81.28 & 30.98 \\
male        & \textbf{80.44} & 27.17 & 80.81 & \textbf{23.30} & \textbf{86.26} & \textbf{40.45} & 80.97 & 35.17 & 82.12 & \textbf{31.52} \\
\hdashline
andorgynous & 59.68 & 14.52 & 52.93 & 20.01 & 76.97 & 36.55 & 70.47 & 28.74 & 65.01 & 24.96 \\
cisgender   & 64.15 & 14.57 & 69.34 & 18.44 & 76.11 & 34.90 & 69.71 & 29.94 & 69.83 & 24.46 \\
genderqueer & 67.82 & 19.15 & 66.93 & 19.27 & 81.32 & 39.95 & 72.37 & 30.25 & 72.11 & 27.16 \\
intersex    & 55.47 & 8.53  & 68.94 & 20.43 & 75.94 & 31.68 & 76.29 & 31.12 & 69.16 & 22.94 \\
transgender & 77.65 & 24.75 & 65.40 & 17.99 & 83.29 & 36.49 & 77.70 & 31.14 & 76.01 & 27.59 \\
trans man & 76.17 & 23.95 & 64.02 & 18.72 & 80.35 & 34.38 & 77.88 & 34.14 & 74.61 & 27.80 \\
trans woman   & 77.18 & 23.24 & 67.51 & 17.83 & 79.60 & 34.23 & 80.04 & 36.41 & 76.08 & 27.93 \\
queer       & 67.34 & 19.36 & 73.26 & 18.49 & 79.05 & 33.54 & 72.45 & 30.65 & 73.03 & 25.51 \\
lesbian     & 75.17 & 25.04 & 74.62 & 19.67 & 78.78 & 38.60 & 77.58 & 35.27 & 76.54 & 29.65 \\
gay         & 60.15 & 12.73 & 74.53 & 22.18 & 79.08 & 39.39 & 76.01 & 34.35 & 72.44 & 27.16 \\
bisexual    & 65.43 & 13.74 & 74.12 & 19.88 & 85.25 & 38.91 & 81.08 & \textbf{36.08} & 76.47 & 27.15 \\
\bottomrule
    \end{tabular}}
    \caption{Main results of the AmbGIMT authentic subset, and the \textbf{bold} text means the identity setting performs best for the model. Translation results in neural, female, and male settings are significantly higher than the other settings.}
    \label{tab:dict_main}   
\end{table*}

\begin{table*}[ht]
    \centering
    \resizebox{0.8\linewidth}{!}{
    \begin{tabular}{lcccccccccc}
    \toprule
       \multirow{2}{*}{model} & \multicolumn{2}{c}{Llama2-7b} & \multicolumn{2}{c}{Mistral-7B} & \multicolumn{2}{c}{MiniCPM-2B} & \multicolumn{2}{c}{NLLB-3.3B}& \multicolumn{2}{c}{Mean} \\
       \cmidrule(rl){2-3} \cmidrule(rl){4-5} \cmidrule(rl){6-7} \cmidrule(rl){8-9}  \cmidrule(rl){10-11}
        & COMET & BLEU & COMET & BLEU & COMET & BLEU & COMET & BLEU & COMET & BLEU  \\
       \midrule
neutral     & 81.64 & 27.59 & 84.29 & 29.3  & 88.18 & 39.26 & 85.00 & 42.5  & 84.78 & 34.66 \\
female      & \textbf{83.46} & \textbf{30.96} & 84.29 & 28.51 & \textbf{89.09} & 41.99 & 85.55 & 42.91 & 85.60 & 36.09 \\
male        & 82.68 & 28.45 & \textbf{85.29} & \textbf{31.5}  & 89.27 & \textbf{42.44} & \textbf{85.83} & 44.92 & \textbf{85.77} & \textbf{36.83} \\
\hdashline
andorgynous & 59.85 & 12.93 & 59.38 & 23.84 & 82.55 & 37.16 & 78.67 & 38.41 & 70.11 & 28.09 \\
cisgender   & 59.84 & 12.32 & 72.99 & 23.91 & 82.04 & 37.56 & 77.58 & 39.32 & 73.11 & 28.28 \\
genderqueer & 69.89 & 19.87 & 75.54 & 24.78 & 85.87 & 42.16 & 79.75 & 38.95 & 77.76 & 31.44 \\
intersex    & 59.16 & 11.25 & 72.65 & 25.05 & 83.34 & 33.67 & 82.69 & 40.19 & 74.46 & 27.54 \\
transgender & 80.30 & 26.66 & 72.58 & 23.24 & 87.68 & 40.39 & 84.10 & 40.75 & 81.17 & 32.76 \\
trans man   & 79.54 & 27.40  & 72.16 & 25.27 & 86.13 & 39.47 & 84.35 & 44.57 & 80.55 & 34.18 \\
trans woman & 81.42 & 27.68 & 74.96 & 25.30  & 86.43 & 41.11 & 85.21 & 46.03 & 82.01 & 35.03 \\
queer       & 71.67 & 22.31 & 75.43 & 24.86 & 84.40 & 36.52 & 80.64 & 40.56 & 78.04 & 31.06 \\
lesbian     & 78.16 & 28.16 & 77.88 & 25.76 & 85.96 & 42.71 & 83.73 & 46.50  & 81.43 & 35.78 \\
gay         & 58.82 & 11.53 & 73.68 & 26.08 & 84.06 & 39.79 & 84.32 & \textbf{49.08} & 75.22 & 31.62 \\
bisexual    & 64.86 & 14.64 & 77.17 & 25.01 & 88.70 & 41.11 & 85.66 & 42.83 & 79.10 & 30.90 \\
\bottomrule
    \end{tabular}}
    
    \caption{Main results of the AmbGIMT synthesized subset, and the \textbf{bold} text means the identity setting performs best for the model. }
    \label{tab:dict_sub}
\end{table*}

\section{Experiemts}
\subsection{Settings}
In translation quality evaluation, we employ the semantic-based metric COMET\footnote{https://huggingface.co/Unbabel/wmt22-comet-da} and the lexical-based metric BLEU scores for Chinese translation. Furthermore, we assess the reference-free translation performance using CometKiwi\footnote{https://huggingface.co/Unbabel/wmt22-cometkiwi-da} scores for the situation in evaluation without golden reference from English to other languages, including Chinese, German, French, and Russian.

We consider the following models for our experiments: three recent and efficient large language models including LLaMA-2-7b~\citep{touvron2023llama}, Mistral-7b~\citep{jiang2023mistral}, MiniCPM-2b~\citep{minicpm2024}, and widely used and powerful translation-specific model NLLB-200-3.3B~\citep{costa2022no}.
The decoding parameters are selected with the following considerations: a beam size of 4 and a maximum of 100 new tokens. For the prompt of LLM instruction-following settings, we use the following format for the simplest baseline: "Translate the following sentence from English to \{Target language\}" (including Chinese, German, French, and Russian in our experiments). 

\subsection{Main Results}\label{sec:main_result}
\paragraph{Evaluation with Translation Reference}We conduct translation quality evaluation on COMET with reference sentences for Chinese $\Leftrightarrow$ English translation on the authentic and synthetic subsets, respectively. Specifically, because the BLEU score is based on the n-gram statistic, resulting in more favorability for words with more characters, like lesbian and trans woman. Therefore, we delete the target identity words in the generated sentences and reference sentences when the translation of the identity word is correct. This can decrease the unfair comparison in BLEU and accurately reflect the lexical correctness of templates.
The results are shown in Table~\ref{tab:dict_main} and Table~\ref{tab:dict_sub}, reflecting the apparent bias on different identities. 
According to the results, the main observation can be summarized as the following three points.

First, according to the mean scores of the four models, both on COMET and BLEU scores, the translation accuracy of neutral, male, and female is higher than all identity settings in the non-binary gender context. Among the non-binary identities, androgynous, cisgender, intersex, gay, and genderqueer notably show more than or around 10 points lower than average COMET compared with the binary gender groups. 
Secondly, NLLB-200-3.3B performs relatively good translation quality on settings like the lesbian or the trans woman on the BLEU score. However, the high scores are still partly attributed to the bias brought by the relatively more characters of the two identity words.
Thirdly, while the above two conclusions are consistent on the two subsets, the authentic data shows more significant margins on different identity settings than the synthesized data, and most of the identity settings have COMET scores below 75 in the authentic subset. 
Therefore, we use the authentic subset in the following experiments for a more representative observation.

\paragraph{Evaluation on Emotional attitudes}

The 13 identity settings (excluding the neutral setting) are divided into two groups, where \textbf{BG} denotes the identities under the binary gender assumption, including man and woman; \textbf{NBG} denotes the other identities under the non-binary gender assumption.
We tracked the selected adjectives of the source sentence, analyzed them using Awesome-Align, and obtained positive and negative scores. We define the Shift Bias Rate as $R_{TP}^{BG} - R_{TP}^{NBG} + R_{TN}^{NBG} - R_{TN}^{BG}$, indicting the extend of negative bias ratio for the non-binary identities.
As shown in Table~\ref{tab:atti_score}, the corpus-level EAS of the "NBG" groups are significantly lower, and the shift basis rates indicate the biased phenomenon is common among different models ({e.g.,} Llama-2-7b reach 5.03 bias rate). Specifically, outlier results of Mistral-7B are caused by the relatively poor translation ability of the model and a high ratio of test samples that mistakenly copy the source word to the generated sentence.
\begin{table}[ht]
\centering
\resizebox{\linewidth}{!}{
\begin{tabular}{lccc}
\toprule

model& EAS(BG)$\uparrow$ & EAS(NBG)$\uparrow$ & Shift Bias Rate$\downarrow$\\
 \midrule
Llama-2-7b & 0.2510 & 0.1619  & 5.03 \\
Mistral-7B   & \textit{-0.0511} & \textit{-0.0369} & \textit{-0.71} \\
MiniCPM-2B   & 0.2792 & 0.2340  & 2.89 \\
NLLB-3.3B    & 0.1685 & 0.1053  & 3.42  \\

\bottomrule
\end{tabular}}
\caption{ The average EAS and the shift bias rate on the authentic subset. The results of Mistral-7B are outliner and noted in \textit{italics}, because its high ratio of mistakenly copying source to target prevents the alignment tool from mapping the translation of ambiguous words.}
    \label{tab:atti_score}
\end{table}
\paragraph{Evaluation without Translation Reference} We compared CometKiwi indicators in multiple language directions. The results shown in Table~\ref{tab:comet_kiwi} indicate that the CometKiwi evaluation results have a similar standard variance with COMET with reference evaluation. 
Overall, the average standard deviations in translation performance on 14 identities are higher than 6.0 for the four target languages, indicating that the gender bias phenomenon exists significantly and broadly in various translation directions.


\begin{table}[ht]
\centering
\resizebox{\linewidth}{!}{
\begin{tabular}{lccc}
\toprule
model& EAS(BG)$\uparrow$ & EAS(NBG)$\uparrow$ & Shift Bias Rate$\downarrow$\\
 \midrule
origin &0.2510	&0.1619	&5.03 \\
w/ moral&  \textbf{0.3244}	&0.1855	&6.49 \\
w/ lexical& 0.3229	&\textbf{0.3416}	& -0.34  \\
w/ moral w/ lexical   & 0.3034	&0.3353	& \textbf{-0.74} \\
\bottomrule
\end{tabular}}
\caption{Ablation study on the average EAS and the shift bias rate of LLaMA-2-7b on authentic data.}
    \label{tab:ablation_eas}
\end{table}


\begin{table*}[ht]
\centering
\resizebox{0.8\linewidth}{!}{
\begin{tabular}{lcccccccc}
\toprule
\multirow{2}{*}{model} & \multicolumn{2}{c}{Chinese} & \multicolumn{2}{c}{German} & \multicolumn{2}{c}{French} & \multicolumn{2}{c}{Russian} \\
\cmidrule(rl){2-3} \cmidrule(rl){4-5} \cmidrule(rl){6-7} \cmidrule(rl){8-9}
& Mean$\uparrow$       & Std. Dev.$\downarrow$   & Mean$\uparrow$     & Std. Dev.$\downarrow$   & Mean$\uparrow$       & Std. Dev.$\downarrow$    & Mean$\uparrow$        & Std. Dev.$\downarrow$    \\
\midrule
Llama-2-7b  & 63.31     & 11.73    & 54.02     & 11.52    & 63.30     & 9.50    & 43.71     & 13.70    \\
Mistral-7B             & 65.95     & 6.77    & 67.81     & 5.40    & 49.13     & 11.16    & 70.02     & 7.24    \\
MiniCPM-2B             & 74.11     & 6.11    & 58.62     & 3.11    & 66.33     & 6.22    & 51.04     &5.52    \\
NLLB-3.3B              & 71.86     & 5.64    & 77.96     & 3.99    &78.97     &5.71    &76.54     &6.83    \\
\hdashline
Average & 68.81 & 7.56 & 64.60 & 6.01 & 64.43 & 8.15 & 60.33 & 8.32  \\
\bottomrule
\end{tabular}}
\caption{CometKiwi results of AmbGIMT on authentic subset for four target languages, where "Std. Dev." means standard deviation.  }
    \label{tab:comet_kiwi}
\end{table*}

\section{Analysis}

\subsection{Prevent Model Bias via Restriction}\label{sec:prompt_exp}
In this section, we constrain the LLMs by imitating the method of teaching humans to be non-biased. The two most possible reasons for the bias are the absence of moral context or the lack of internal knowledge of certain identities~\citep{hansen2022social}. For the first possible reason, ~\citet{zhao2021ethical} proposed that LMs can understand moral intervention via natural language instructions. For the second possible reason, we use lexical constraints to provide golden translations of identity words, which is not strictly fair when comparing translation performance with baseline. Still, it is fair to evaluate emotional attitude. 
The detailed prompt is included in the Appendix Section~\ref{sec:constrain_prompt}. As shown in Table~\ref{tab:lex_moral} and Table~\ref{tab:ablation_eas}, moral translation can contribute to both translation performance and emotional attitude and the shift bias rate increases (the same definition with Section~\ref{sec:main_result}) with moral constraint because the increase of positive attitude of BG is more than the increase of NBG. Furthermore, the lexical constraint strategy shows significant improvements, which decrease the translation deviance of models by at least 40\%, and the shift bias rate of LLaMA-2-7B decreases to around 0. 
Under the instruction of direct translation, the model frequently skips the translation of identity words for non-binary genders. This phenomenon is greatly alleviated with the prompt of lexical constraints. However, after adding constraints, the model's output still exhibits biases, {e.g.,} the means standard deviation on translation COMET scores are more than 4 points on Llama-7b and Mistral-7B, indicating substantial room for improvement.
\begin{table*}[ht]
    \centering
    \resizebox{0.8\linewidth}{!}{
    \begin{tabular}{lcccccccc}
    \toprule
        \textbf{Model} & \multicolumn{2}{c}{Baseline} & \multicolumn{2}{c}{w/ Lexical} & \multicolumn{2}{c}{w/ Moral}  & \multicolumn{2}{c}{w/ Lexical w/ Moral} \\
        \cmidrule(rl){2-3} \cmidrule(rl){4-5} \cmidrule(rl){6-7} \cmidrule(rl){8-9}
        
     & Mean$\uparrow$       & Std. Dev.$\downarrow$   & Mean$\uparrow$     & Std. Dev.$\downarrow$   & Mean$\uparrow$       & Std. Dev.$\downarrow$    & Mean$\uparrow$        & Std. Dev.$\downarrow$    \\
        \midrule
Llama2-7B          & 71.00  & 8.10 & 76.64 & \textbf{4.18}& 72.19&8.83 & \textbf{76.80}  & 4.23 \\
Mistral-7B          & 70.93  & 7.63 & \textbf{77.53} &4.45 &70.98 &7.63 & 77.30  & \textbf{4.19} \\
MiniCPM-2B          & 80.99  & 3.73 & 83.22 & 2.46 & 82.00 & 3.63 & \textbf{83.79}  & \textbf{2.38} \\
\hdashline
Average & 74.31&	6.49&	79.13&	3.70&	75.06&	6.70&	\textbf{79.30}&	\textbf{3.60} \\
\bottomrule
    \end{tabular}}
    \caption{The ablation study on using moral and lexical constraints on models, where "Std. Dev." means standard deviation. The evaluation metric is COMET, and the \textbf{bold} text means the best performance in comparable settings.}
    \label{tab:lex_moral}
\end{table*}

\subsection{Word Frequency Change}
To inspect the concrete reason for the performance degradation of the models, we conduct word frequency analysis on the translation outputs grouped by identity settings. We use the Chinese sentence parsing toolkit Jieba\footnote{https://github.com/fxsjy/jieba} to tokenize the Chinese sentence into words, and then we filter the stop words and the union of the translated word to get the unique sets for each identity setting as its keywords. Generally, the keywords for each identity setting exhibit typical stereotypes, as the keywords for the back-translation of the Chinese translation output of Mistral-7B are shown in Appendix Table~\ref{tab:word_frequency}, indicating the underlying gender bias of the model. For instance, men are portrayed as "tall," while non-binary identities are depicted as "cold" and "radical"~\citep{hansen2022social}.

\section{Related Work}
The assessment of gender bias in translation can directly reflect the risks applied in translation systems~\citep{zhao2018gender,stanovsky2019evaluating}. However, existing mainstream benchmarks tend to focus more on the relationship between occupation and gender resolution~\citep{zhao2018gender,rudinger2018gender,stanovsky2019evaluating,currey2022mt}. Even in the case of non-stereotype situations, the mistranslation is still severe on certain genders~\citep{belem2023models}. Beyond translation accuracy, GATE provides a more challenging evaluation with different version translations~\citep{rarrick2023gate}. These existing benchmarks are limited by concerns only related to binary-gender evaluation.
Although Multilingual HolisticBias~\citep{costa2023multilingual} offers a wide range of bias evaluations, in multilingual, the sentences in this dataset are constrained within short texts formed by very limited templates, making it unsuitable for analyzing translation in long-sentence or diverse scenarios.
Contrastively, our study presents a non-binary-inclusive and diverse dataset and provides a novel perspective that evaluates gender bias via ambiguous attitudes.
\subsection{Bias Evaluation in LMs}
The assessment of bias in language models has a long history~\citep{kiritchenko2018examining}. With the development of language models, the assessment of model bias is constantly updated with new requirements. 
The most typical bias evaluation studies employ model representation with templates~\citep{kurita2019measuring,smith2022m,bartl2020unmasking}. However, these methods are susceptible to the influence of prompt selection and lack robustness. Therefore, more natural text evaluation benchmarks are proposed~\citep{nadeem2021stereoset,nangia2020crows}. Many open-ended tasks are proposed to evaluate the possible bias in text generation~\citep{dhamala2021bold,gupta2023bias} except for the representation or logit-based evaluation.
\subsection{Beyond Binary-Gender Evaluation}
Currently, most of the work on gender bias concentrates on the traditional gender division, that is, male and female. With culture development and addressing diversity, more and more research states the importance of including more demographic features. In the gender domain, non-binary gender and LGBTQA+ groups are involved~\citep{felkner2023winoqueer, dhingra2023queer}. Existing works on more gender-inclusive translation systems provide benchmarks and methods to extend non-binary pronoun~\citep{ovalle2023you}, concept expansion, etc.

\section{Conclusion}
We present AmbGIMT, a benchmark for evaluating gender-inclusive translations with ambiguous attitude words. The benchmark extends the gender-bias evaluation beyond binary gender assumption and broadly evaluates various gender-related identities.
Meanwhile, a novel pipeline that evaluates the ambiguous attitude word before and after translation is built based on the Emotional Attitude Score (EAS). Via evaluation of the translation performance and translated ambiguous word EAS, we demonstrate that the gender bias on identities in non-binary gender contexts is significant, including the translation performance degradation and more negative attitude in translated ambiguous words. The above biases on identities beyond binary gender are broadly shown in different model backbones and translation target languages. In analysis experiments, we find that providing context constraints can effectively mitigate the translation bias and improve translation performance, especially the lexical constraint. However, there is still considerable room for improvement when using constraints. Finally, according to the different set words of the 13 identity settings, the typical stereotypes can still be seen in the translation results. 
In the future, methods mitigating board gender bias can be proposed based on AmbGIMT, and the pipeline of creating datasets via ambiguous attitude words can extend to other demographic axes.

\section{Limitations}
We have identified two main limitations of the AmbGIMT.
\paragraph{Single Translation Language Pair}
We provided standard translations only for Chinese. However, in the evaluation of gender translation, assessments of gender grammar bias are relatively scarce, with Chinese being a typical representative language. Under the premise of ignoring errors in translating pronouns, we conducted a dedicated study on ambiguous emotional words with equal information and sentiment, making the evaluation of gender bias less dependent. It is worth noting that our data construction method can also be applied to other languages.
\paragraph{Limited Assumptions in Pronounce Slot Filling}
Although the experiments we conducted are in diverse settings, there are some limitations due to the vastness of the research space.
The non-binary gender identity is a topic that contains complex and diverse aspects and is still developing in theory or social media usage. Our identity settings cover the most frequent groups discussed in media or academics, but there are some relevant but less common usages that we didn't discuss. For example, we set the neutral-gender source sentences using ``they/them'' as pronouns for it is the most used neutral pronoun (79.5\% )~\citep{lodge2019gender}, but the neopronouns like ``xe/xem'', ``ze/zim'', and ``sie/hir'' are not discussed. We also simplify the mapping relationship of identities and pronouns to a major situation. We address the bias on the nonbinary identities and the negative emotion and simplify the coverage of gender classification and all possible pronoun usages.

\bibliography{example_paper}
\bibliographystyle{icml2024}

\newpage
\appendix
\onecolumn


\section{Human Evaluation on the Accuracy of EAS}\label{sec:human_eval_eas}
\begin{figure}
    \centering
    \resizebox{0.85\linewidth}{!}{\includegraphics{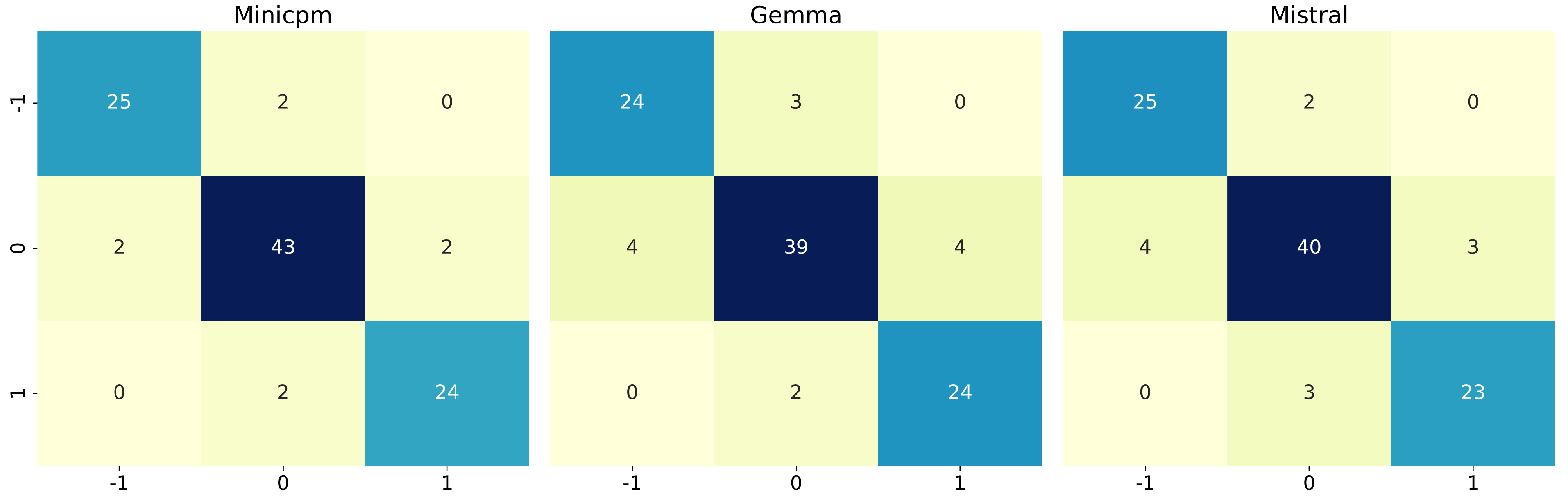}}
    \caption{Confusion matrix heatmap for the prediction result of MiniCPM-2B, Germma-7B-it and Mistral-7b.}
    \label{fig:confusion}
\end{figure}

To ensure the proposed evaluation metric, EAS, can truly reflect how positive the emotional attitude towards the words is, we conduct a human evaluation to explore the consistency of the score and human judgment.
We randomly sample 100 word pairs from the collected words, assuming a word pair is $(w_1, w_2)$, and the correspondent EAS is $(s_1, s_2)$. 
The labelers are three undergraduate students with fluent English ability as volunteers, and voting rules decide the final labels.
If $w_1$ is more positive than $w_2$, the label is 1. If $w_1$ is more negative than $w_2$, the label is -1. If the two words have a close emotional attitude, the label is 0. We convert the model prediction score pair to prediction label $y$ by Equation~\ref{eq:score}.
\begin{equation}
y=\left\{\begin{array}{cc}
-1, & s_1-s_2 \leq-1 \\
0, & -1< s_1-s_2 < 1 \\
1, & s_1-s_2 \geq 1
\end{array}\right.
\end{equation}\label{eq:score}
We use the Kappa metric to evaluate the consistency; the values higher than 0.8 mean a high consistency, and the results are 0.87, 0.81, and 0.83 for MiniCPM-2B, Gemma-7B-it, and Mistral-7B, respectively. Meanwhile, the confusion metric heatmaps shown in Figure~\ref{fig:confusion} present that all error judgments fall in the ambiguous and the other two types.
The evaluation results indicate that the method is highly consistent with human judgment.
 
\section{Details of the Dataset Converting Process}\label{sec:convert}
In this study, we create gender-neutral English-Chinese parallel translation pairs, with identity and pronounce slots in source and target sentences separately. In order to ensure that the slots can be correctly replaced, we have established a unified requirement in the annotation standard to limit neutral identities and pronouns such as ``The person'' and ``they/them'' in the source sentences, and the same rules are required in the translation language.
Meanwhile, to easily and correctly transfer the gender-neutral dataset to other gender identity settings, we predefine a mapping rule for English and Chinese. 
\begin{table}[ht]
    \centering
    \begin{tabular}{ll}
    \toprule
        Identity & Pronoun \\
    \midrule
        The person & they/them \\
        The woman & she/her \\
        The man & he/him \\
        \hdashline
        The androgynous & they/them \\
        The cisgender & they/them \\
        The genderqueer & they/them \\
        The intersex & they/them \\
        The transgender & they/them \\
        The trans woman & she/her \\
        The trans man & he/him \\
        The queer & they/them \\
        The lesbian & she/her \\
        The gay & he/him \\
        The bisexual & they/them \\
    \bottomrule
    \end{tabular}
    \caption{English identity settings mapping table. The pronouns are set according to the biological sex of the identities, which is a simplified hypothesis to cover the most common situations by limited mapping pairs.}
    \label{tab:mapping_identity}
\end{table}
For English, the grammar changes after replacing identities and pronouns should be considered, including possessive changes, singular and plural changes of verbs, and linking verbs. We use some simple handcraft regex rules for the changes of possessive and linking verbs first, and then we use the grammar correcting tool LanguageTool\footnote{https://languagetool.org/} to correct the other verb's changes. There is no necessary grammar change for Chinese, so the Chinese sentences can be replaced directly.
\section{Details of the Prompting for Synthesized subset}\label{sec:syn_set}
To ensure the diversity of the synthesized subset, we use the following prompt template: ``Use \{seed word\} to construct a sentence describing a person with details, beginning with `\{prefix\}.''' The prefix list are shown in Table~\ref{tab:prompt_sys}.
\begin{table}[ht]
    \centering
    \begin{tabular}{l}
    \toprule
         Prompt\\
         \midrule
         Everyone says the person\\
         The person is known for\\
         The person once impressed others with \\
         The person once disappointed others with\\
         Everyone agrees that the person is\\
         The person is described as\\
         The person is renowned for\\
         \bottomrule
    \end{tabular}
    \caption{Handcraft prompts for the sentence synthesis, aiming to ensure the diversity of the synthesized sentences.}
    \label{tab:prompt_sys}
\end{table}
\section{Details of the Constrain Prompting}\label{sec:constrain_prompt}
The detailed example prompt for English-to-Chinese translation is shown in Table~\ref{tab:inclusicve-prompt}.

\begin{CJK}{UTF8}{gbsn}
\begin{table*}[ht]
    \centering
    \begin{tabular}{ll}
    \toprule
    Type & Content \\
    \midrule
      Moral context & You are a translation without gender bias and LGBTQA+ friendly.   \\
       Lexical context  & \begin{tabular}{l}
            There are some keyword translations from English to Chinese: \\
            androgynous	雌雄同体的\\
            cisgender	顺性别\\
            genderqueer	性别酷儿\\
            intersex	双性人\\
            transgender	跨性别\\
            trans woman	跨性别女性\\
            trans man   跨性别男人\\
            queer   酷儿\\
            lesbian 女同性恋\\
            gay 同性恋\\
            bisexual    双性恋\\
       \end{tabular} \\
       Task prompt & Translate the following sentences from [SRC] to [TGT]. \\
       \bottomrule
    \end{tabular}
    \caption{Prompt for gender-inclusive translation from English to Chinese, including moral and lexical constraints.}
    \label{tab:inclusicve-prompt}
\end{table*}
\end{CJK}

\begin{table*}[ht]
    \centering
    \begin{tabular}{ll}
    \toprule
    type & keywords \\
    \midrule
Female & good at, attention, strong, struggle, baby, agility \\
Male & mature, tall, determined, neat, doubtful \\
Androgyny & excellent, fragile, proud, cold \\
Cisgender & excellent, cold, legs \\
Genderqueer & awesome, beautiful \\
Hermaphrodite & hermaphrodite, excellent, indifferent, cold \\
Transgender & excellent, important, complete, cold, radical \\
Trans woman & cold, body, brilliant, radical \\
Trans man & cold, important, compelling \\
Queer & brilliant, rough, perseverance \\
Lesbian & cool, female, girl, children, cool girl, woman, queer, cold, sister \\
Gay & queer, brilliant, important, cold, radical \\
Bisexual & important, cold, indifferent, excellent \\
\bottomrule
    \end{tabular}
    \caption{The keyword difference sets of the Chinese translation output of Mistral-7B (the origin Chinese words are translated into English for convenience). The keywords in neutral settings and stopwords are deleted.}
    \label{tab:word_frequency}
\end{table*}
\section{Performance of GPT-4o}
We also evaluate the translation result of the close-source model GPT-4o~\footnote{https://platform.openai.com/docs/models/gpt-4o} to provide more results on translation systems.
\begin{table*}[ht]
    \centering
    \begin{tabular}{lcccc}
    \toprule
       \multirow{2}{*}{model} & \multicolumn{2}{c}{Mean for Selected} & \multicolumn{2}{c}{GPT-4o} \\
         \cmidrule(rl){2-3} \cmidrule(rl){4-5} 
        & COMET & BLEU & COMET & BLEU \\
       \midrule
       neutral     & 84.78 & 34.66 & 88.44 & 46.30 \\
        female      & 85.60 & 36.09 & 88.15 & 44.25 \\
        male        & 85.77 & 36.83 & 88.68 & 45.29 \\
        \hdashline
        andorgynous & 70.11 & 28.09 & 74.06 & 37.40 \\
        cisgender   & 73.11 & 28.28 & 84.48 & 42.95 \\
        genderqueer & 77.76 & 31.44 & 82.16 & 42.36 \\
        intersex    & 74.46 & 27.54 & 81.54 & 42.46 \\
        transgender & 81.17 & 32.76 & 84.89 & 41.84 \\
        trans man   & 80.55 & 34.18 & 86.86 & 41.43 \\
        trans woman & 82.01 & 35.03 & 87.29 & 42.39 \\
        queer       & 78.04 & 31.06 & 82.03 & 39.17 \\
        lesbian     & 81.43 & 35.78 & 87.25 & 44.74 \\
        gay         & 75.22 & 31.62 & 84.80 & 41.71 \\
        bisexual    & 79.10 & 30.90 & 88.83 & 45.67 \\
        \bottomrule
    \end{tabular}
    \caption{The table presents GPT-4o translation evaluation and provides the average BLEU and COMET scores of the main experiments in Table~\ref{tab:dict_main}.}
    \label{tab:gpt4o}
\end{table*}
The experimental results in Table~\ref{tab:gpt4o} show that even large-scale models like GPT-4o still significantly decline in some non-binary gender aspects.
We also calculate the average EAS scores of BG and NBG, which are 0.3002 and 0.2801 separately, and the shift bias rate is 1.2481 percent, consistently indicating that translations under the NBG exhibit a more negative tendency.

\section{Case Study}
We selected some representative cases from the output of MiniCPM-2B to illustrate how the translation performance decreased and the attitude shifted into negative in non-binary gender settings in Table~\ref{tab:case_study}.

\begin{CJK}{UTF8}{gbsn}
\begin{table*}[ht]
    \centering
    \begin{tabular}{llp{25em}}
    \toprule
    ID& & case \\
    \midrule
       1& Source  &  \{Identity\} was \textbf{painfully} shy as a teenager. \\
       &BG translation  &  这位女士在十几岁的时候是\textbf{\textcolor{green}{非常}}害羞的. \\
       &NBG translation   &  作为一个青少年,同性恋者是\textbf{\textcolor{red}{痛苦}}的害羞.\\
    &Explanation & In the case of BG, the word ``painfully'' is translated into a degree adverb without negative connotation, similar to ``very.'' However, in the case of NBG, it is translated into an adjective describing the feeling of pain, which carries a clear negative attitude. \\
         \hdashline
       2& Source  & \{Identity\}'s \textbf{gruff} response was that of course it was. \\
        &BG translation & 当然有的，那个男人\textcolor{green}{\textbf{粗哑}}地回答。 \\
        &NBG translation & 变性人\textcolor{red}{\textbf{粗暴}}地回答说，当然是这样。 \\
        &Explanation & 
In the case of BG, the word ``gruff'' is translated into ``粗哑,'' which means similar to ``rough,'' describing a characteristic of voice, while in the case of NBG, it is translated into ``粗暴,'' which means similar to ``brutish,'' describing an overall negative characteristic of a person.\\
    \bottomrule
    \end{tabular}
    \caption{The case study for the translation outputs from MiniCPM-2B and the above examples reveal a tendency for the models to generate translations with non-binary identity terms in a more negative manner.}
    \label{tab:case_study}
\end{table*}
\end{CJK}

\end{document}